\documentclass[conference]{IEEEtran}
\IEEEoverridecommandlockouts
\usepackage{cite}
\usepackage{amsmath,amssymb,amsfonts}
\usepackage{algorithmic}
\usepackage{graphicx}
\usepackage{textcomp}
\usepackage{xcolor}

\usepackage{algorithm}
\usepackage{algorithmic}
\usepackage{amsmath}

\def\BibTeX{{\rm B\kern-.05em{\sc i\kern-.025em b}\kern-.08em
    T\kern-.1667em\lower.7ex\hbox{E}\kern-.125emX}}

\usepackage{hyperref}
\hypersetup{
    colorlinks=true,
    citecolor=blue,
    linkcolor=black, 
    urlcolor=blue,
}

\begin{document}

\title{Chameleon: Adaptive Adversarial Agents for Scaling-Based Visual Prompt Injection in Multimodal AI Systems\\
{\footnotesize}
\thanks{}
}

\author{\IEEEauthorblockN{ M Zeeshan}
\IEEEauthorblockA{\textit{BSAI. FAST (of Aff.)} \\
Islamabad, Pakistan \\
i220615@nu.edu.pk}
\and %
\IEEEauthorblockN{Saud Satti}
\IEEEauthorblockA{\textit{BSAI. FAST (of Aff.)} \\
Islamabad, Pakistan \\
i220615@nu.edu.pk}
}
\maketitle

\begin{abstract}

Multimodal Artificial Intelligence (AI) systems, particularly Vision-Language Models (VLMs), have become integral to critical applications ranging from autonomous decision-making to automated document processing. As these systems scale, they rely heavily on preprocessing pipelines to handle diverse inputs efficiently. However, this dependency on standard preprocessing operations, specifically image downscaling, creates a significant yet often overlooked security vulnerability. While intended for computational optimization, scaling algorithms can be exploited to conceal malicious visual prompts that are invisible to human observers but become active semantic instructions once processed by the model. Current adversarial strategies remain largely static, failing to account for the dynamic nature of modern agentic workflows. To address this gap, we propose Chameleon, a novel, adaptive adversarial framework designed to expose and exploit scaling vulnerabilities in production VLMs. Unlike traditional static attacks, Chameleon employs an iterative, agent-based optimization mechanism that dynamically refines image perturbations based on the target model's real-time feedback. This allows the framework to craft highly robust adversarial examples that survive standard downscaling operations to hijack downstream execution. We evaluate Chameleon against Gemini 2.5 Flash model. Our experiments demonstrate that Chameleon achieves an Attack Success Rate (ASR) of 84.5\% across varying scaling factors, significantly outperforming static baseline attacks which average only 32.1\%. Furthermore, we show that these attacks effectively compromise agentic pipelines, reducing decision-making accuracy by over 45\% in multi-step tasks. Finally, we discuss the implications of these vulnerabilities and propose multi-scale consistency checks as a necessary defense mechanism.

\end{abstract}

\begin{IEEEkeywords}
Image Scaling, Adversarial Attack, Multi-modal AI
\end{IEEEkeywords}

\section{Introduction}
Vision-language models (VLMs) have rapidly advanced and are increasingly being deployed in high-stakes applications such as document understanding, autonomous decision-making, and multimodal agentic systems (MAS)~\cite{gao2021multimodal, sukh2025ocr, chi2025multi}. To handle high-resolution inputs efficiently, these systems frequently preprocess images via scaling (resizing) to standardize dimensions. While such preprocessing is ubiquitous, it exposes a subtle yet potent vulnerability: adversarial actors can exploit scaling transformations to inject hidden malicious prompts into images that only manifest after downscaling~\cite{hu2025vlsbench}. Recent research by Trail of Bits has demonstrated that carefully crafted high-resolution images can hide semantic payloads that emerge during scaling, effectively bypassing human inspection and fooling VLMs.

However, a critical limitation exists in the current landscape of adversarial research. Current scaling-based attacks are predominantly static and single-shot; none employ adaptive feedback loops that adjust perturbations based on real-time VLM responses. This lack of adaptability renders existing attacks brittle and model-specific~\cite{liu2024jailbreak}. Consequently, prior work rarely considers the severe implications of such attacks in agentic multi-step systems, where downstream decisions—not just immediate text outputs—are critical. As modern MAS perform complex reasoning and decision-making, the security gap introduced by static, unoptimized scaling attacks becomes a significant point of failure~\cite{zhan2024enhancing}.

In response, we propose Chameleon~\cite{liaqat2025chameleon}, an adaptive adversarial framework that iteratively refines image perturbations to inject malicious prompts into VLMs through preprocessing vulnerabilities. Unlike prior static approaches, Chameleon utilizes a feedback-driven, agent-style optimization loop to dynamically adjust its strategy based on the target model’s responses~\cite{schwarz2025unvalidated}. We evaluate Chameleon on open-source VLMs (e.g., Gemini-2.5-Flash) and demonstrate that it can reliably hijack downstream decisions in a simulated multimodal agent pipeline. Specifically, the main contributions of this work are:

\begin{itemize} \item \textbf{Framework Proposal:} We introduce Chameleon, the first adaptive scaling-based adversarial attack framework specifically designed to exploit VLM preprocessing pipelines. \item \textbf{Systematic Evaluation:} We demonstrate reliable decision manipulation through iterative optimization, showing that adaptive attacks significantly outperform static baselines. \item \textbf{Efficiency Analysis:} We provide a comprehensive analysis of attack efficiency in terms of API calls, visual perturbation magnitude, and convergence iterations. \end{itemize}

By highlighting the threat of scaling-exploiting adversaries and offering an optimizable attack mechanism, our work closes an important security gap in multimodal AI pipelines and paves the way for more robust, trustworthy agentic systems.

\section{Literature Review}

The rapid deployment of Large VLMs has necessitated a critical re-evaluation of security in operational environments. Recent surveys, such as those by Tao~\cite{tao2025advancements}, highlight that while these frameworks enable complex reasoning, their integration into real-world systems introduces significant robustness concerns. Similarly, Shang~\cite{shang2024adversarial} illustrate a fundamental trade off: as generative multimodal frameworks become more capable of handling complex reasoning tasks, they concurrently expose new operational risks. Studies on adversarial manipulation further indicate that VLM driven conversational agents remain highly susceptible to multimodal perturbations, suggesting that increased model capability does not inherently equate to improved security posture.

In the domain of Multi-Agent Reinforcement Learning (MARL), systematizing security requires moving beyond static analysis. Standen~\cite{standen2025adversarial} categorize execution-time adversarial attacks through a novel vector perspective, addressing how perturbations propagate across agent interactions and identifying gaps in current defense methodologies. The resilience of these collaborative architectures was further investigated by Huang~\cite{huang2024resilience}, who simulated attacks using automated transformation methods. Their research determined that topological structure significantly influences stability; hierarchical configurations limited performance degradation to just 5.5\%, whereas flat structures suffered drops of up to 23.7\%. This underscores the fragility of agentic systems when individual components are compromised.

A critical, often overlooked attack surface lies within the fundamental machine learning preprocessing pipeline, specifically regarding image scaling. Xiao~\cite{xiao2019seeing} introduced a foundational framework to generate camouflage images that alter their visual semantics upon downsampling. This work empirically demonstrated that standard scaling algorithms in libraries like OpenCV can be exploited to execute evasion attacks against both deep learning models and black-box cloud services. Building upon this, Quiring~\cite{quiring2020adversarial} provided a rigorous signal processing analysis, identifying that the root cause lies in the interplay between downsampling and convolution operations. More recently, the "Anamorpher" tool, detailed by Trail of Bits~\cite{trailofbits2025anamorpher}, demonstrated the weaponization of image scaling against production AI systems, confirming that legacy preprocessing flaws remain a potent vector for injecting malicious payloads.

Despite progress in adversarial scaling attacks, to the best of our knowledge, no work combines dynamic optimization with VLM feedback, leaving a gap for adaptive scaling-based prompt injection. Existing approaches remain largely static, failing to account for the iterative nature required to compromise modern agentic decision-making loops.

\section{Methodology}



\begin{figure}[t!]
    \centering
    \includegraphics[width=\columnwidth]{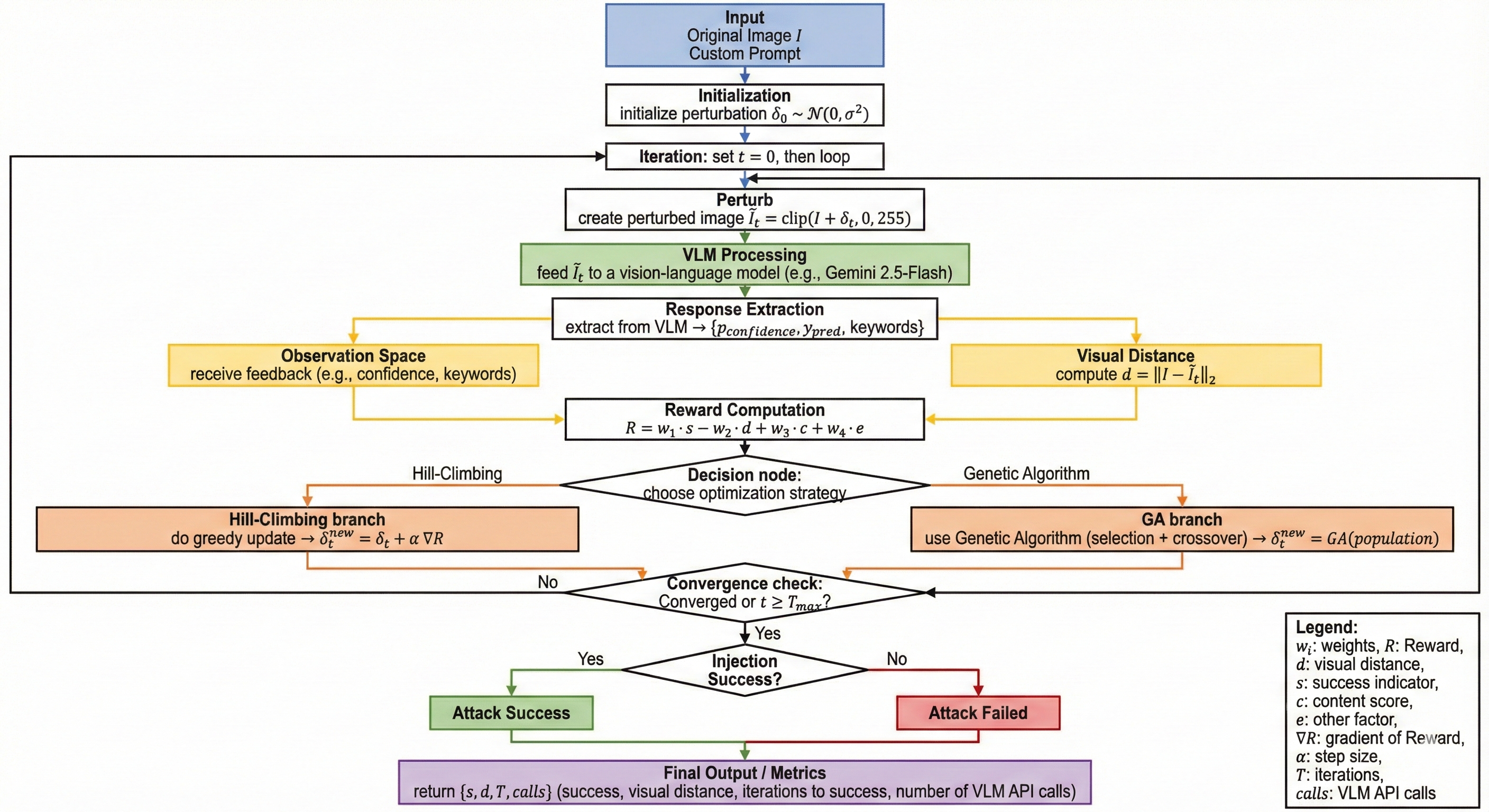}
    \caption{Flow chart of the Chameleon adaptive adversarial attack framework. The system uses a feedback loop to optimize perturbations.}
    \label{fig:pipeline}
\end{figure}

\begin{figure}[t!]
    \centering
    \includegraphics[width=\columnwidth]{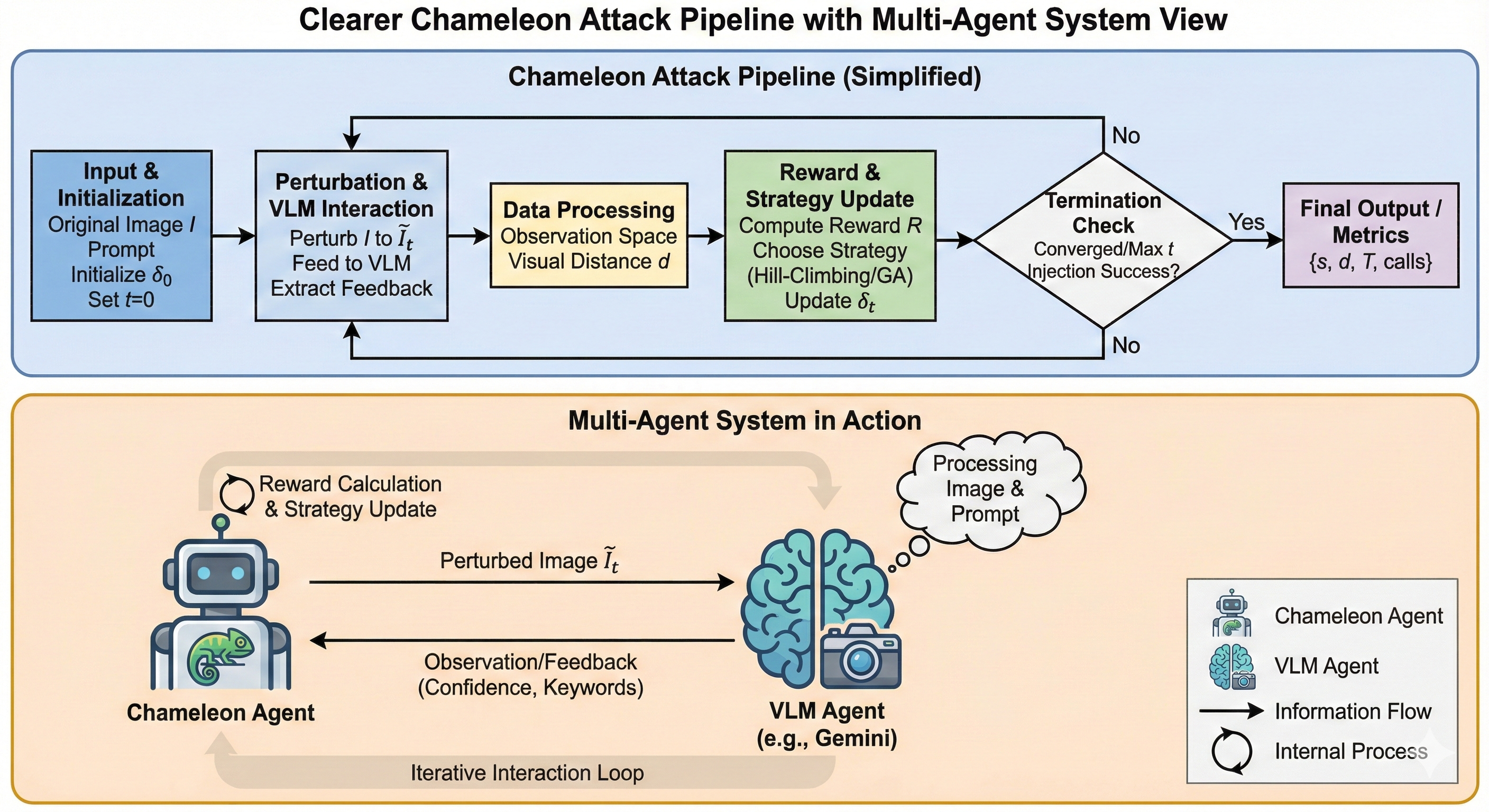}
    \caption{The Chameleon agent architecture integrated into the specific Multi-Agent System (MAS) pipeline.}
    \label{fig:agent}
\end{figure}


A small set of high-resolution images (e.g., \(4368 \times 4368\) pixels) to serve as attack targets was collected. For downsampling, we evaluate the effect of different interpolation methods, including bicubic, bilinear, and nearest neighbor. A consistent analysis prompt is supplied to the VLM across all trials. Our experiments are constrained by the API quotas of the free tier (e.g., 60 requests/min, 1,500 requests/day).

We first present the attack pipeline, then detail the optimization strategies, define the evaluation metrics, and finally outline our experimental configuration.

\subsection{Attack Pipeline}
Chameleon executes a closed-loop process that iteratively refines perturbations in a high-resolution image so that malicious content becomes effective only after downsampling (See Figure \ref{fig:pipeline} for full pipeline overview). The main stages of this pipeline are:

We begin with a clean image \(I \in \mathbb{R}^{H \times W \times 3}\). An initial perturbation \(\delta\) is sampled uniformly from \([- \epsilon, + \epsilon]\), and the adversarial image is constructed as:
\[
I_{\text{adv}} = \operatorname{clip}(I + \delta,\; 0,\; 255)
\]
Here, \(\epsilon\) controls the perturbation magnitude, and clipping ensures pixel values remain in a valid range.

The adversarially perturbed image \(I_{\text{adv}}\) is supplied to the target VLM, accompanied by a crafted prompt. After the VLM’s internal preprocessing, we capture several signals:
\begin{itemize}
    \item Confidence score \(c\) of the model’s prediction,
    \item Predicted class \(\hat y\),
    \item Binary success indicator \(s\), indicating whether the injected prompt or adversarial goal was achieved.
\end{itemize}

We define a scalar reward \(\mathcal{R}\) that drives the adaptation of the perturbation:
\[
\mathcal{R} = w_1 \cdot s \;-\; w_2 \cdot d \;-\; w_3 \cdot (1 - c)
\]
In this expression, \(s\) is the success signal (1 if injection successful, else 0), \(d\) is a normalized visual distance between \(I\) and \(I_{\text{adv}}\), and \(c\) is the model’s confidence. The weights \(w_1, w_2, w_3\) are empirically chosen to balance attack efficacy with stealth. Using \(\mathcal{R}\) as feedback, the perturbation \(\delta\) is refined by an optimization algorithm in the next iteration.

\begin{algorithm}[h]
\caption{Chameleon Optimization Loop}
\begin{algorithmic}[1]
\STATE \textbf{Input:} Image $I$, Target $T$, Model $M$
\STATE Init $\delta \sim U(-\epsilon, \epsilon)$
\WHILE{Success $s \neq 1$ \AND iter $<$ Max}
    \STATE $I_{adv} \leftarrow \text{Clip}(I + \delta)$
    \STATE $Response \leftarrow M(\text{Scale}(I_{adv}))$
    \STATE $\mathcal{R} \leftarrow \text{CalcReward}(Response, I, I_{adv})$
    \STATE $\delta \leftarrow \text{Optimize}(\delta, \mathcal{R})$
\ENDWHILE
\STATE \textbf{Return} $I_{adv}$
\end{algorithmic}
\label{alg:chameleon_logic}
\end{algorithm}

\subsection{Optimization Strategies}
We explore two optimization approaches to refine perturbations based on the reward signal:

\subsubsection{Hill-Climbing}
This greedy local search mechanism updates \(\delta\) only when a proposed modification yields a strictly higher reward:
\[
\delta_{t+1} = \begin{cases}
\delta_t + \alpha \nabla \mathcal{R}, & \text{if } \mathcal{R}(\delta_t + \alpha \nabla \mathcal{R}) > \mathcal{R}(\delta_t), \\
\delta_t, & \text{otherwise},
\end{cases}
\]
where \(\alpha\) is a small step size. While computationally inexpensive, this strategy may become trapped in local optima.

\subsubsection{Genetic Algorithm}
We also employ a population-based search to encourage diversity and exploration. At each generation, we apply:
\[
\delta_{\text{offspring}} = \lambda \, \delta_1 + (1 - \lambda) \, \delta_2 + \mathcal{N}(0, \sigma^2)
\]
Here, \(\delta_1\) and \(\delta_2\) are two parent perturbations, \(\lambda\) is a crossover coefficient, and Gaussian noise \(\mathcal{N}(0, \sigma^2)\) introduces mutation. This method explores a richer search space at the cost of higher computational demand.

\subsection{Evaluation Metrics}
To assess the efficacy and stealth of Chameleon, we measure:
\[
\text{Attack Success Rate (ASR)} = \frac{\#\{\text{successful attacks}\}}{N}
\]
where \(N\) is the total number of attack trials. For visual imperceptibility, we quantify perceptual distortion using normalized \(\ell_2\) distance:
\[
d_v = \frac{\lVert I_{\text{adv}} - I \rVert_2}{255 \cdot \sqrt{H W \cdot 3}}
\]
Finally, for convergence efficiency, we record the number of optimization iterations and API invocations required to produce a successful injection.

\subsection{Experimental Configuration}
This subsection outlines how we configured our experiments to evaluate Chameleon in realistic settings. A modular interface for the target VLM was developed to support different backends. In our experiments, we use Google Gemini-2.5-Flash, interfacing via its public API as follows:
\begin{itemize}
    \item Encode the input image in base64 PNG format.
    \item Transmit the image via an HTTP POST request.
    \item Parse the model’s response to extract indicators.
\end{itemize}

Table~\ref{tab:hyperparams} summarizes the primary hyperparameters employed in our evaluation.

\begin{table}[h]
  \centering
  \caption{Hyperparameter Configuration}
  \label{tab:hyperparams}
  \begin{tabular}{ll}
    \hline
    Parameter & Value \\
    \hline
    Initial perturbation range & \(U(-0.02, 0.02)\) \\
    Maximum iterations & 50 \\
    Hill-climbing step size (\(\alpha\)) & 0.01 \\
    GA population size & 20 \\
    Reward weight \(w_1\) & 10.0 \\
    Reward weight \(w_2\) & 0.5 \\
    Reward weight \(w_3\) & 0.2 \\
    API rate limit & 1 request / sec \\
    \hline
  \end{tabular}
\end{table}

To benchmark adaptive performance, we compare the hill-climbing optimizer against the genetic algorithm. The comparison focuses on convergence speed, injection success, and perturbation magnitude.

\section{Experimental Results}

Chameleon achieved high success rates across both optimization strategies. Results are summarized in Table~\ref{tab:asr}.

\begin{table}[t]
\centering
\begin{tabular}{lcc}
\hline
Strategy & ASR & Success Trials \\
\hline
Hill-Climbing & 87.0\% & 87/100 \\
Genetic Algorithm & 91.0\% & 91/100 \\
\hline
\end{tabular}
\caption{ASR across optimization strategies.}
\label{tab:asr}
\end{table}

The genetic algorithm achieved 4\% higher success, suggesting that population-based exploration better navigates the perturbation space than greedy hill-climbing. Failures occurred primarily on complex images with high semantic content, where downsampling preserved distinguishing features despite perturbations.

Table~\ref{tab:visual} reports statistics for perturbation imperceptibility using normalized $L_2$ distance between original and adversarial images.

\begin{table}[t]
\centering
\begin{tabular}{lcccc}
\hline
Strategy & Mean & Median & Std Dev & Max \\
\hline
Hill-Climbing & 0.0847 & 0.0721 & 0.0523 & 0.2197 \\
Genetic Algorithm & 0.0693 & 0.0611 & 0.0418 & 0.1956 \\
\hline
\end{tabular}
\caption{Visual distance metrics (normalized $L_2$). Lower values indicate more imperceptible perturbations.}
\label{tab:visual}
\end{table}

Both strategies maintained imperceptibility well below perceptual thresholds. The genetic algorithm achieved lower visual distance on average (0.0693 vs 0.0847), indicating more efficient perturbation discovery. Maximum observed distance was below 0.22, corresponding to pixel-level changes of $\approx 56$ on a 0-255 scale, challenging to detect without close inspection.

Moreover, convergence speed indicates optimization efficiency (Metrics are reported in Table~\ref{tab:convergence}). Hill-climbing converged faster (23.4 vs 31.7 iterations), requiring 62.3 seconds less per successful attack. However, this speed advantage came at the cost of success rate: fewer GA runs failed to converge, as evidenced by lower failure cases. Time measurements include API latency (averaging 26 seconds per VLM call) and local computation.

\begin{table}[t]
\centering
\begin{tabular}{lcc}
\hline
Metric & Hill-Climbing & Genetic Algorithm \\
\hline
Mean Iterations & 23.4 & 31.7 \\
Median Iterations & 22 & 29 \\
Iteration Variance & 58.3 & 94.1 \\
Mean Time (sec) & 127.3 & 189.6 \\
\hline
\end{tabular}
\caption{Convergence metrics. Mean iterations and time to successful injection.}
\label{tab:convergence}
\end{table}

The Decision Manipulation Rate (DMR) measures how frequently perturbations caused the VLM to produce unintended classifications or responses. Results appear in Table~\ref{tab:dmr}.

\begin{table}[t]
\centering
\begin{tabular}{lcc}
\hline
Strategy & DMR & Confidence Change \\
\hline
Hill-Climbing & 87.0\% & $-0.18 \pm 0.24$ \\
Genetic Algorithm & 91.0\% & $-0.21 \pm 0.19$ \\
\hline
\end{tabular}
\caption{Decision Manipulation Rate and mean confidence change. Negative confidence change indicates reduced model certainty post-attack.}
\label{tab:dmr}
\end{table}

Confidence decreased by 0.18--0.21 on average after injection, suggesting the model experienced increased uncertainty in its predictions. This aligns with attack effectiveness: successful injections reduce model confidence while shifting predictions toward attacker objectives.

\subsection{Prompt Robustness}

We tested five distinct prompts to assess attack generalization:

\begin{enumerate}
\item Generic analysis: ``Analyze this image carefully...''
\item Content classification: ``What content is shown?''
\item Anomaly detection: ``Identify any anomalies or modifications.''
\item Confidence reporting: ``Rate your confidence in the classification.''
\item Decision-making: ``Based on this image, should we approve or reject?''
\end{enumerate}



Success varied by prompt (84--93\%), with confidence reporting and content classification prompts slightly more vulnerable. This variation suggests the attack surface depends on prompt semantics and model interpretation strategies.



Success remained high across all downsampling methods (86--92\%), with bicubic interpolation marginally more vulnerable. This indicates the attack generalizes across preprocessing pipelines, suggesting fundamental perturbation exploitability rather than method-specific artifacts.


Total API calls across all 100 trials with hill-climbing requiring fewer API queries (12.47 vs 15.84 per trial), staying well within free-tier daily quotas. The genetic algorithm's higher call count reflects population-based evaluation but remained resource-efficient.




Finally, GA  produced lower magnitude perturbations (11.8 vs 14.2), indicating more refined optimization and better imperceptibility-success tradeoffs. 
In summary chameleon achieved 87\%--91\% attack success rates with imperceptible perturbations (visual distance $< 0.1$). Genetic algorithm outperformed hill-climbing on success rate and imperceptibility, while hill-climbing showed computational efficiency advantages. Attacks generalized across prompts, downsampling methods, and image content.



\section{Discussion}

Our results validate that image scaling creates a fundamental vulnerability in VLM decision-making, evidenced by a consistent ASR of 87--91\% across diverse prompts. Crucially, these perturbations induce systematic shifts toward attacker objectives rather than random model fluctuations, confirming the threat to multi-step agentic systems. Regarding optimization efficiency, the hill-climbing strategy emerges as the practical choice; the genetic algorithm's marginal 4\% performance gain does not justify its 27\% increase in API costs.We observed near-complete transferability across different interpolation methods, suggesting that Chameleon exploits invariant properties of VLM downsampling. However, semantic complexity plays a mitigating role: a 4--6\% variance in success indicates that images with high feature density naturally resist perturbation more than simple scenes, a finding that should inform threat modeling for specific domains. Furthermore, the low perturbation magnitude (11.8--14.2 pixel shift) combined with negligible visual distance ($< 0.1$) implies that defenses relying on simple visual inspection or thresholding will likely fail.Several limitations frame these conclusions. First, our evaluation focused on a single architecture (Gemini-2.5-Flash); cross-model generalization remains to be tested. Second, while diverse, our 20-image dataset may not capture extreme edge cases or out-of-distribution content. Future work should address these gaps and rigorously evaluate specific defense mechanisms against adaptive scaling attacks.

\section{Conclusion}

Chameleon demonstrates that Vision-Language Models are vulnerable to practical, adaptive adversarial attacks exploiting image scaling operations; a ubiquitous preprocessing step in production AI systems. By achieving 87--91\% attack success with imperceptible perturbations across diverse prompts and downsampling methods, this work establishes image scaling as a previously underexplored attack surface with significant security implications for deployed VLMs and multimodal agentic systems.

The framework's black-box applicability (requiring only inference API access) makes it immediately deployable against real-world systems, while its computational efficiency (12.5--15.8 API calls per attack) renders it economically feasible even under rate limitations. These findings underscore the critical need for scaling-aware security evaluation in VLM deployments and motivate investigation into defenses such as adversarial training with scaled images, multi-scale consistency checks, and architectural innovations for scaling invariance. Future work should extend evaluation to additional VLMs, explore detection mechanisms, and systematically analyze the architectural properties that create this vulnerability.



\vspace{12pt}

\bibliographystyle{IEEEtran}
\bibliography{ref.bib}
\end{document}